\documentclass[a4paper,twoside]{article}

\usepackage{epsfig}
\usepackage{subcaption}
\usepackage{calc}
\usepackage{amssymb}
\usepackage{amstext}
\usepackage{amsmath}
\usepackage{amsthm}
\usepackage{multicol}
\usepackage{pslatex}
\usepackage{apalike}
\usepackage{algorithm2e}
\usepackage[bottom]{footmisc}

\RequirePackage{booktabs}
\usepackage{hyphenat}
\usepackage[table, dvipsnames]{xcolor}
\usepackage{siunitx}
\usepackage{makecell}

\usepackage{pifont}
\newcommand{\cmark}{\ding{51}}%
\newcommand{\xmark}{\ding{55}}%

\usepackage{array}
\newcolumntype{H}{>{\setbox0=\hbox\bgroup}c<{\egroup}@{}}

\usepackage{SCITEPRESS}     

\begin{document}

\title{FutrTrack: A Camera-LiDAR Fusion Transformer for 3D Multiple Object Tracking}
\author{\authorname{Martha Teiko Teye\sup{1}\orcidAuthor{0000-0002-2370-4700}, Ori Maoz\orcidAuthor{0000-0001-5661-830X} 
and Matthias Rottmann\sup{2}\orcidAuthor{0000-0003-3840-0184}}
\affiliation{\sup{1}University of Wuppertal, Germany}
\affiliation{\sup{2}Institute of Computer Science, Osnabrück University, Germany}
\email{m.teye-hk@uni-wuppertal.de, matthias.rottmann@uos.de}
}

\keywords{Multi-modal Transformer Tracking, Camera–LiDAR Fusion, 3D Multi-Object Tracking}

\abstract{We propose FutrTrack, a modular camera-LiDAR multi-object tracking framework that builds on existing 3D detectors by introducing a transformer-based smoother and a fusion-driven tracker. Inspired by the query-based tracking frameworks, FutrTrack employs a multi-modal two-stage transformer refinement and tracking pipeline. Our fusion tracker integrates bounding boxes with multi-modal bird’s eye view (BEV) fusion features from multiple cameras and LiDAR without the need for an explicit motion model.
The tracker assigns and propagates identities across frames, leveraging both geometry and semantic cues for robust re-identification under occlusion and viewpoint changes. Prior to tracking, we refine sequences of bounding boxes with a temporal smoother over a moving window to refine trajectories, reduce jitter, and improve spatial consistency. 
Evaluated on nuScenes and KITTI, FutrTrack demonstrates how query-based transformer tracking methods benefit greatly from multi-modal sensor features compared to previous single sensor modality approaches. With an aMOTA of 74.7 on the nuScenes test set, FutrTrack achieves strong performance on 3D MOT benchmarks, significantly reducing identity switches while maintaining competitive accuracy. Our approach offers an efficient framework to improve transformer-based trackers to compete with other neural network-based trackers even with limited data and without pre-training.}

\onecolumn \maketitle \normalsize \setcounter{footnote}{0} \vfill

\section{\uppercase{Introduction}}

Multi-Object Tracking (MOT) aims to simultaneously detect and maintain consistent identities of multiple targets over time, forming a fundamental component of perception systems in areas such as autonomous driving, robotics, and video surveillance~\cite{mot20}. Traditional 2D MOT methods operate on image sequences, relying heavily on appearance and motion cues to associate detections frame by frame. While these approaches have achieved remarkable progress with advances in deep learning, their performance is often limited by occlusions, depth ambiguity, and the lack of precise spatial understanding inherent in 2D imagery. \begin{figure}
    \centering
    \includegraphics[width=0.48\textwidth]{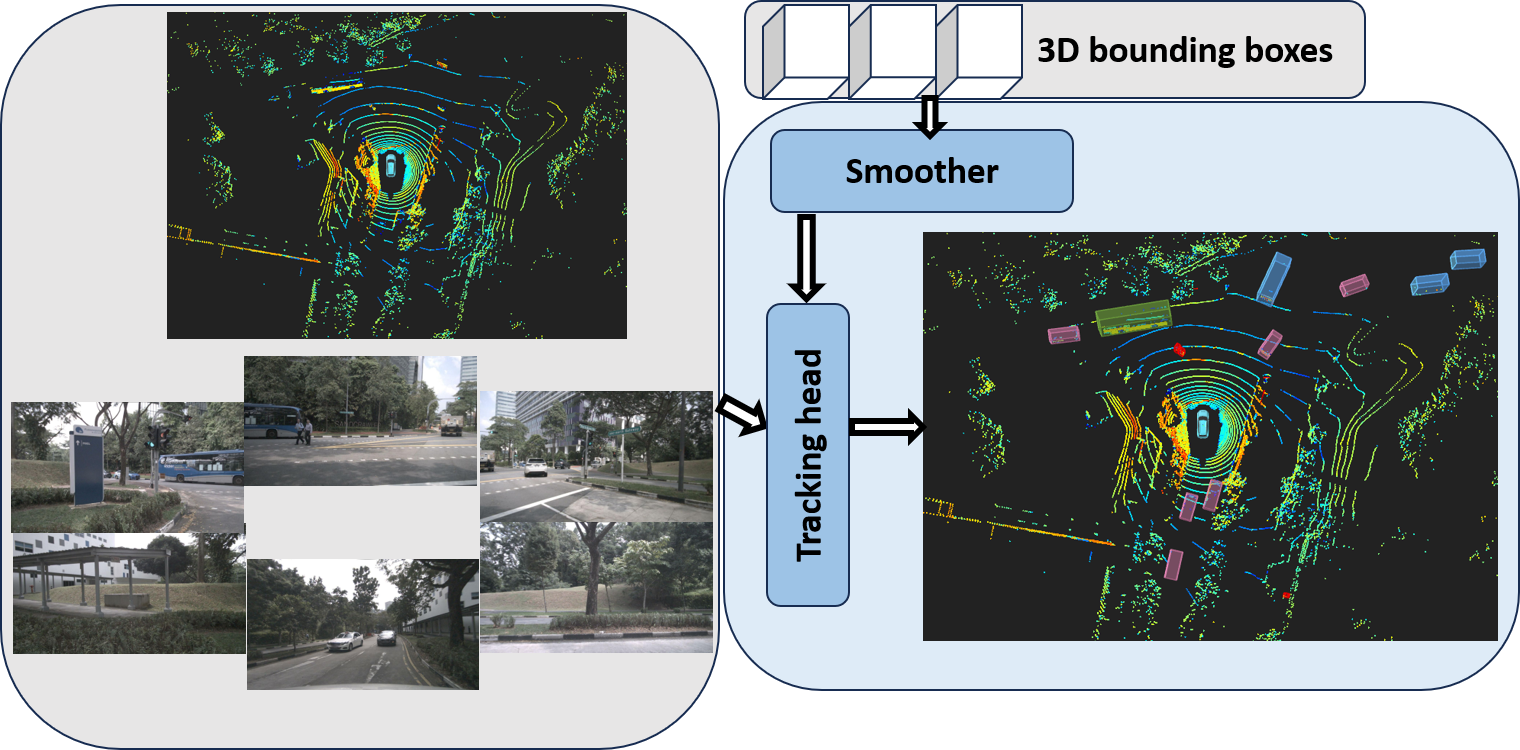}
    \caption{FutrTrack extracts LiDAR and multi-view camera features and tracks them using query-based fusion attention mechanism. It also refines bounding boxes predictions from existing detectors through a smoother model to serve as queries for the fusion-based attention transformer.}
    \label{fig:enter-label}
\end{figure} 
To address these challenges, research has increasingly shifted towards 3D Multi-Object Tracking~\cite{simpletrack,sort}, which leverages geometric information from sensors such as LiDAR or stereo cameras to provide accurate localization and motion estimation in real-world coordinates. By incorporating the third spatial dimension, 3D MOT enables more reliable tracking in dynamic environments, forming a crucial step toward robust perception in autonomous systems. However, these methods can be noisy, and even methods that attempt full end-to-end tracking~\cite{Wang2024MCTrackAU,10755965,meinhardt2022trackformermultiobjecttrackingtransformers} find it difficult to optimize because they are tightly coupled to specific detectors making identity assignment a harder downstream task.

\begin{figure*}

    \centering
    \includegraphics[width=0.99\textwidth]{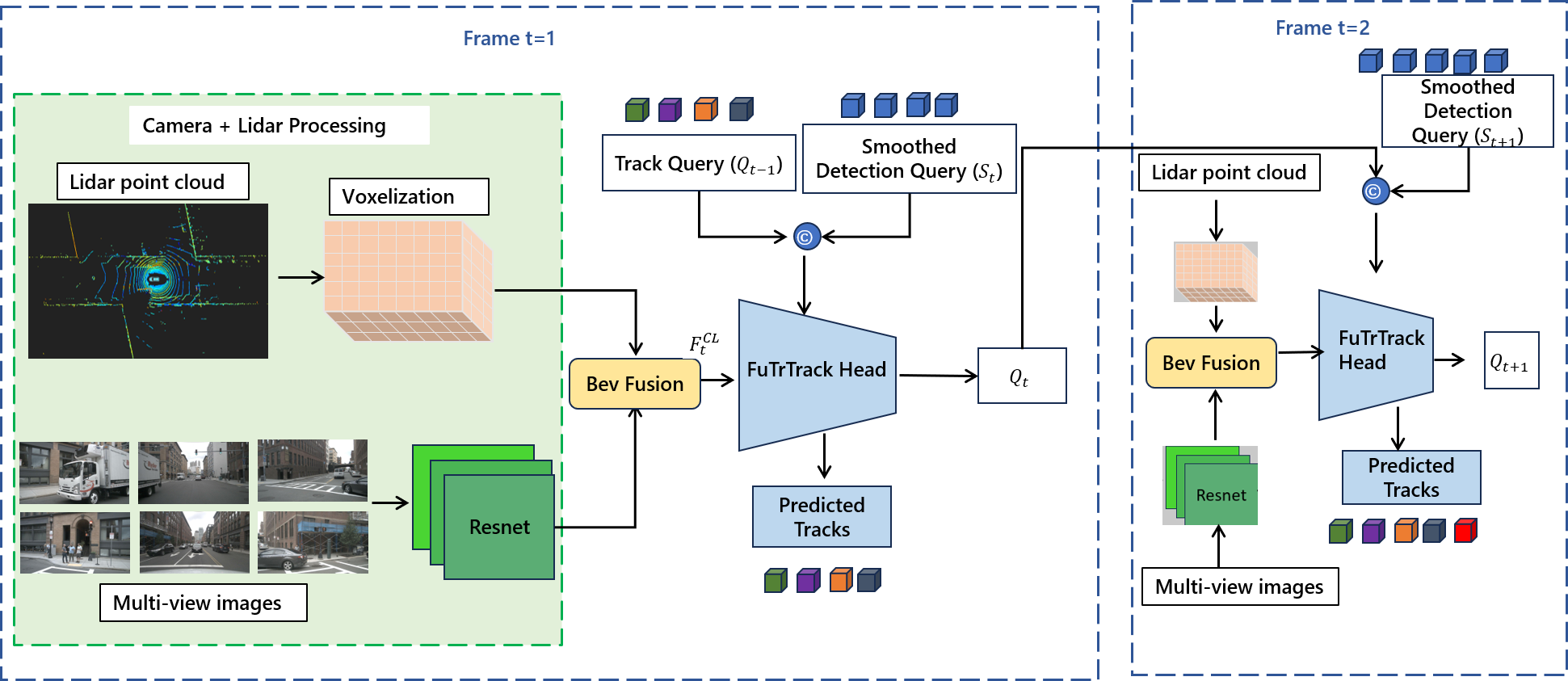}
    \caption{Overall architecture of FutrTrack. Features from camera and LiDAR are extracted  using a standard 2D ResNet and voxel-based encoders respectively. The multi-modal features are projected into a birds-eye-view (BEV) space to produce unified fused features. The fused features together with detection queries and track queries from previous frames are processed by the FutrTrack to produce tracks.}
    \label{fig:overall_architecture}
\end{figure*}
Most recently, transformer query-based tracking has been used for both object detection and tracking tasks. 
Transformer query-based features represent scene elements as a set of learnable queries that interact with encoded sensor data through attention mechanisms, allowing the network to model object relationships in a structured way. In the camera domain, query-based frameworks have achieved strong performance for object detection, yet their effectiveness for object tracking remains limited, largely due to difficulties in maintaining temporal consistency across frames. Recent studies show that combining multiple sensing modalities, particularly camera and LiDAR, can significantly enhance query-based object detection by providing complementary spatial and semantic cues~\cite{focalformer3d,10160968bevfusion}, but this multi-modal fusion paradigm has not been sufficiently explored in the context of transformer-based multi-object tracking.
In detection task, these query-based methods have produced state-of-the-art results and proven that fusing LiDAR and camera features in BEV space provides strong semantic and geometric cues for re-identification. However, most existing approaches to query-based tracking have not yet consistently achieved state-of-the-art (SOTA) performance. These query-based tracking methods~\cite{motr_camera,Mutr3d_camera} are also mostly implemented for 2D tracking, but this has not been explored enough for 3D tracking. 
This motivates a modular framework where detection refinement and identity tracking using transformers are treated as complementary stages.


We propose FutrTrack, a multi-modal tracker that harnesses a holistic camera-LiDAR fusion approach to provide rich attention queries for track association. FutrTrack makes use of existing object detectors, refines them using a smoother model, and then tracks them using our new multi-modal query-based tracker that relies on multi-modal features extracted from both camera and LiDAR sensors. The query-based tracker when evaluated on standard benchmark shows improved average Multi Object Tracking Precision (aMOTP) with smoother trajectories and reduced identity switches on the nuScenes~\cite{caesar2020nuscenes} and KITTI~\cite{kitti2012CVPR} datasets.

Our main contributions in this paper are as follows:
\begin{itemize}
   \item A multi-modal transformer-based tracking framework that extends query-based tracking to jointly reason over camera and LiDAR modalities, enabling robust multi-object tracking in fused 3D space.
   \item A camera–LiDAR projection and fusion module that aligns image features with LiDAR point representations to construct a unified bird’s-eye-view (BEV) embedding optimized for tracking rather than detection.
    \item On nuScenes and KITTI, our method outperforms prior transformer-based trackers, particularly in occlusion scenarios and for small objects.

\end{itemize}

\section{\uppercase{Related Work}} \label{sec:work}
MOT has seen significant developments in recent years, particularly with the incorporation of deep learning methods~\cite{lidarreview,li3detr}.
Current MOT research generally falls into one of the following paradigms: Tracking-by-Detection~\cite{deepsort,sort,Wang2024MCTrackAU}, Joint Detection and Tracking~\cite{meinhardt2022trackformermultiobjecttrackingtransformers,TWB_2d}, and Tracking-by-Attention~\cite{motr_camera,Mutr3d_camera,3dMotformer}. Methods in the Joint Detection and Tracking and Tracking-by-Attention categories largely depend on rich object feature extraction, which demands substantial GPU computation and data. Such requirements are difficult to meet with the limited processing capacity of autonomous vehicles, especially in online modes. In addition, these approaches often do not outperform the simpler Tracking-by-Detection strategy in practice. For these reasons, this work focuses on a lightweight transformer Tracking-by-Attention method.

Traditional MOT pipelines separate object detection and tracking, i.e. detection first, then data association like in SORT~\cite{sort}, DeepSORT~\cite{deepsort} and extended Kalman filtering (EKF). With transformer-based detectors like DETR~\cite{detrE2E}, end-to-end tracking approaches emerged, using track queries across frames to fuse detection with temporal continuity~\cite{meinhardt2022trackformermultiobjecttrackingtransformers,motr_camera,Zhang2023MotionTrackET}. MOTR~\cite{motr_camera} adapts that paradigm for image-based queries, embedding spatial queries, and enabling joint detection and tracking directly from camera data.

Baseline methods such as CenterPoint-Track~\cite{Centerpoint,simpletrack}, perform data associations using Kalman Filter, Extended Kalman Filter, and other probabilistic models. Prior fusion tracking works mostly follow detection-first pipelines, but few integrate fusion features within end-to-end MOT frameworks. Recently, a DETR-based MOT, LiDAR MOT-DETR~\cite{teye2025lidarmotdetrlidarbasedtwostage} exploits point-based or voxel-based features to localize and track objects in 3D. This method is particularly good at refining bounding boxes, but is limited to only considering the LiDAR-based feature when matching objects to tracks.
\paragraph{Multi-Modal (LiDAR–Camera) Fusion for Detection and Tracking:}
Fusion methods range from early fusion (raw-level concatenation), through feature-level fusion (e.g., PointPainting~\cite{vora2020pointpaintingsequentialfusion3d}, PointAugmenting~\cite{li2020pointaugmentautoaugmentationframeworkpoint}), to BEV-level fusion (e.g. BEVFusion~\cite{10160968bevfusion}, TransFusion~\cite{bai2022transfusionrobustlidarcamerafusion}). BEVFusion explicitly projects camera and LiDAR features into BEV, enabling transformer-based or convolutional backbones to reason jointly. These have proven to work well in object detection tasks, and FutrTrack exploits these applications in query-based tracking. 


\paragraph{Appearance for Re-identification:}
Camera-based tracking benefits significantly from appearance features. ReID modules (e.g., in DeepSORT~\cite{deepsort}) use image embedding distances to enforce identity consistency. Introducing appearance cues into LiDAR-based or fusion-based trackers remains under-explored, especially in unified DETR-style models. This gap is what our work seeks to bridge and serve as a baseline to camera-LiDAR fusion tracking using transformers.

Our method sits at the intersection of DETR-based MOT, BEV-level multi-modal fusion, and appearance-aware tracking. FutrTrack is, to the best of our knowledge, the first framework to integrate all three.


\section{\uppercase{Method}}
FutrTrack is a query-based transformer multi-object tracker inspired by LiDAR MOT-DETR~\cite{teye2025lidarmotdetrlidarbasedtwostage} by introducing a camera-LiDAR fusion step and extending query-based tracking to work with these fused features. The overall architecture and components are described in figure \ref{fig:overall_architecture}.
With two main components, smoother $m$ and fusion tracker $\Phi$, FutrTrack refines the predictions of the bounding boxes and tracks them using a query-based fusion attention mechanism.
FutrTrack works by selecting bounding boxes with centers $x,y,z$, size $w,l,h$, yaw angle $\theta$ and confidence $c$ from an object detector. A set of these bounding boxes $B = \{x,y,z,w,l,h,\theta,c\}$, across a temporal window $\tau$ are refined using a similar smoothing mechanism $m$ as LiDAR MOT-DETR.
\begin{equation}
    \mathcal{S}_t = m(\mathcal{D}_\tau), 
\end{equation}
where, $\mathcal{D}_{\tau} ={B_1, B2, \ldots, B_N}$. The smoothed bounding boxes $\mathcal{S}_t$, then serve as regions of interest used as queries for the transformer tracker $n$.
The tracker $n$ accepts fused features $\mathcal{F}^\mathit{CL}$ from the camera $\mathcal{F}_c$ and LiDAR $\mathcal{F}^L$ BEV features, as well as query embeddings $Q$ from the smoother.
The FutrTrack feature extraction process consists of two modal encoders; a voxel-based LiDAR encoder and a standard 2D ResNet image encoder.
Then a BEV fusion backbone projects image features into BEV and fuses with LiDAR features via cross-attention or concatenation.
Our transformer detection-tracking head uses queries that carry identity and spatial tokens, predicting bounding boxes and track identities.
Our training objective is to combine detection loss (classification + localization), tracking loss (Hungarian matching across frames), and optionally appearance-reID loss.


\paragraph{LiDAR and Image Encoders:}
The point clouds are encoded into BEV voxel features. A Voxelnet \cite{voxelnet} backbone processes these into high-level BEV feature maps.
Then each of the six camera views is processed via a shared CNN acting as a 2D backbone such as ResNet-50~\cite{he2015resnet} or Swin-Transformer~\cite{liu2021swintransformerhierarchicalvision}, to produce multi-scale feature maps. These feature maps are projected from the 2D view into the 3D ego coordinates to obtain the BEV camera features using camera calibration and learnt attention in a manner similar to BEVFusion~\cite{qi2024ocbev}.

\paragraph{BEV Fusion Backbone:}
The LiDAR and image BEV feature maps are fused using concatenation followed by convolution. The LiDAR and image BEV maps are stacked in a channel-wise manner and processed with convolutional layers. This is followed by transformer-style cross-attention to align LiDAR BEV query features with image BEV key/value features. This fused BEV representation now contains aligned geometric and appearance-rich semantics.

\paragraph{Transformer Query-Based Tracking:}
The proposed tracker is based on a transformer framework and leverages the early fusion of LiDAR and camera features. FutrTrack adopts query-based tracking methods~\cite{motr_camera,Mutr3d_camera,Zhang2023MotionTrackET} where the smoothed detection queries are generated from embeddings of refined bounding boxes from the smoother model.
Then it uses a set of object queries (i.e. tracks) as tracking queries. Each track query is initialized either from scratch (for new objects) or by propagating previous frame queries for track continuity.
Queries attend to fused BEV features via multi-head cross-attention, decoding into box predictions (location, size, orientation), class scores, and identity embeddings.
The track attention mechanism incorporates appearance-aware tokens from the BEV features. The fused features from camera and LiDAR modalities, pooled along the projected object region from the bounding boxes, are injected into query embeddings to enrich re-identification and produce final tracks, $\mathcal{T}$.


\paragraph{Track Query Initialization:} During query initialization at the start of every data sequence, all detections are initialized into queries. 
Via cross-attention, FutrTrack projects these queries into the fused camera-LiDAR features. 
In DETR~\cite{detrE2E}, the self-attention mechanism primarily serves to suppress redundant predictions. Our approach extends this design by introducing the attention layer, which explicitly incorporates interactions across frames. As depicted in Fig. \ref{fig:overall_architecture}, track queries are constructed by concatenating the previous tracked object queries with the detected object queries generated for the current frame. These hybrid queries act as both keys and values within the multi-head attention module. 

Following LiDAR MOT-DETR, object queries are represented as embeddings from bounding boxes of existing detectors. To exploit spatial reasoning inherent in sequential frame data, FutrTrack recurrently makes use of queries from the previous frame that are propagated into the current frame. In this implementation, the track queries from the most recent frame are merged with current detection queries to form the input set.

 At each time step $t$, the inputs to the tracker are:
\begin{itemize}
  \item \textbf{Track queries}, denoted $Q_{t-1}$, which represent active track queries propagated from the previous frame.
  \item \textbf{Refined detections}, $\mathcal{S}_{t}$, obtained from the smoother first-stage.
  \item \textbf{Multi-modal context features}, $\mathcal{F}^\mathit{CL}_{t}$, formed by early fusion of LiDAR and camera encodings.
\end{itemize}

The tracker produces the set of confirmed tracks at time $t$:
\begin{equation}
  \mathcal{T}_{t} = \Phi \!\left(Q_{t-1}, \mathcal{D}_{t}, \mathcal{F}^\mathit{CL}_{t}\right)
\end{equation}

\paragraph{Feature Fusion and Encoding:} 
Each LiDAR sweep and the corresponding image are encoded separately to obtain $\mathcal{F}^{L}$ and $\mathcal{F}^\mathit{C}$. These are combined at an early stage using concatenation or attention-based fusion to yield a joint embedding $\mathcal{F}^\mathit{CL} \in \mathbb{R}^{d}$. The fused embeddings from the most recent frames are aggregated to provide spatio-temporal context. 

\paragraph{Transformer Encoding:} 
The propagated track queries and detection queries are embedded independently:
\begin{align}
  \mathbf{Q}_{t-1} &= \mathrm{Enc}_{Q}\!\left(\mathcal{Q}_{t-1}\right) \\
  \mathbf{S}_{t} &= \mathrm{Enc}_{S}\!\left(\mathcal{S}_{t}\right)
\end{align}
The multi-head attention module then integrates these with the fused history:
\begin{equation}
  \mathbf{Z}_{t} = \mathrm{MHA}\!\left(Q=\mathbf{Q}_{t-1},\; K=\mathbf{S}_{t},\; V=\mathcal{F}^\mathit{CL}_{t}\right)
\end{equation}

\begin{table*}[t]
\caption{Comparison between FutrTrack and LiDAR MOT-DETR on  nuScenes and KITTI validation sets. We evaluate both methods using the same smoother trained on the same multi-modal detector while varying the tracker modality. DM=detector modality,  TM=tracker modality.}
\label{tracker-robust}
\centering
(a) Nuscenes Dataset
\resizebox{\textwidth}{!}{%
\begin{tabular}
{
  l 
  l  
  c  
  c  
  c  
  c  
  c  
  c  
  c  
  c  
  c  
  c  
  c  
  c  
  c  
}
\toprule
\multicolumn{1}{c}{\textbf{Method}} &
\multicolumn{1}{c}{\textbf{Detector}} &
\multicolumn{1}{c}{\textbf{DM}} &
\multicolumn{1}{c}{\textbf{TM}} &
\multicolumn{8}{c}{\textbf{aMOTA} $\uparrow$} &
\multicolumn{1}{c}{\textbf{aMOTP} $\downarrow$} &
\multicolumn{1}{c}{\textbf{IDS} $\downarrow$} &
\multicolumn{1}{c}{\textbf{FRAG} $\downarrow$} \\
\cmidrule(lr){5-12}
\multicolumn{4}{c}{} &
\multicolumn{1}{c}{Overall} &
\multicolumn{1}{c}{Bic.} &
\multicolumn{1}{c}{Bus} &
\multicolumn{1}{c}{Car} &
\multicolumn{1}{c}{Mot.} &
\multicolumn{1}{c}{Ped.} &
\multicolumn{1}{c}{Tra.} &
\multicolumn{1}{c}{Tru.} &
\multicolumn{1}{c}{} &
\multicolumn{1}{c}{} &
\multicolumn{1}{c}{} \\
\midrule

LiDAR MOT-DETR&FocalFormer3D&L               & L &0.736 & 0.622
&0.857 &0.825 &0.777 &0.735 &0.487 &0.708&0.532 &392&468             \\
LiDAR MOT-DETR&FocalFormer3D&C+L               & L &0.738 & 0.636
&\underline{0.860} &0.829 &\underline{0.798} & 0.738&0.497&0.712 & 0.508&399&464              \\
FutrTrack (Ours) &FocalFormer3D &C+L  &C+L  & \textbf{0.759}      &\textbf{0.687} & 0.857&\textbf{0.856} &\textbf{0.827} & \textbf{0.839}& \textbf{0.541} & \textbf{0.725}& \textbf{0.460} &340&429      \\
         \midrule
FutrTrack (Ours)&SparseFusion&C+L                & L & 0.698& 0.540
&0.853 & 0.836 & 0.660& 0.736&0.514 & 0.688&   0.549   &559&516          \\
FutrTrack (Ours) &SparseFusion  &C+L &C+L  &  0.732     & 0.613 
& \textbf{0.861}& \underline{0.845}&0.731 & 0.798&0.517 &0.692 &  0.525 &547&490    \\
\bottomrule

\end{tabular}
}

(b) KITTI Dataset\\
\resizebox{0.7\textwidth}{!}{%
\begin{tabular}
{
  l 
  l  
  c  
  c  
  c  
  c  
  c  
  c  
  c  
  c  
}
\toprule
\multicolumn{1}{c}{\textbf{Method}} &
\multicolumn{1}{c}{\textbf{Detector}} &
\multicolumn{1}{c}{\textbf{DM}} &
\multicolumn{1}{c}{\textbf{TM}} &
\multicolumn{3}{c}{\textbf{Car} $\uparrow$} &
\multicolumn{3}{|c}{\textbf{Pedestrian} $\uparrow$} \\
\cmidrule(lr){5-7}
\cmidrule(lr){7-10}
\multicolumn{4}{c}{} &
\multicolumn{1}{c}{HOTA} &
\multicolumn{1}{c}{MOTA} &
\multicolumn{1}{c}{MOTP} &
\multicolumn{1}{c}{HOTA} &
\multicolumn{1}{c}{MOTA} &
\multicolumn{1}{c}{MOTP} \\
\midrule

LiDAR MOT-DETR&FocalFormer3D&L               & L &0.852 & 0.913 & 0.793&0.642& 0.781 &0.599
\\
LiDAR MOT-DETR&FocalFormer3D&C+L           &L  &0.903&0.901  & 0.823 &0.673 &0.787&0.612

\\
FutrTrack (Ours) &FocalFormer3D &C+L  &C+L  &\textbf{0.946}   &  \textbf{0.915}    &\textbf{0.856} &\textbf{0.711}&\textbf{0.806} &\textbf{0.650}
\\
\bottomrule

\end{tabular}
}
\end{table*}

\paragraph{Track Decoding.} 
The resulting representation $\mathbf{Z}_{t}$ is passed through a decoder network $\mathrm{Dec}(\cdot)$ to generate the final track outputs:
\begin{equation}
  \mathcal{T}_{t} = \mathrm{Dec}(\mathbf{Z}_{t})
\end{equation}
with N objects where each object candidate in $\mathcal{T}_{t}$ consists of an estimated object state 
\[
\{(x_j, y_j, z_j, l_j, w_j, h_j, \theta_j, c_j, \mathrm{id}_j)\}_{j=1}^{N_t}
\]
including position, size, orientation, confidence $c_j$, and identity label $\mathrm{id}_j$.
Tracks with confidence below a threshold $\delta$ are discarded. Surviving tracks are retained as part of $\mathcal{Q}_{t}$ for the next time step, thereby enabling temporal continuity.

\paragraph{Training Loss.} 
A bipartite matching strategy aligns predicted tracks $\mathcal{T}_{t}$ with ground truth $\mathcal{G}_{t}$. Let $\Pi$ denotes the set of valid assignments such that $\pi \in \{1,2,3, \ldots, N\} \mapsto \{1,2,3, \ldots, N\}$ The training objective is:
\begin{equation}
  \mathcal{L}_{\mathrm{total}} = \min_{\pi \in \Pi} \sum_{(j,k)\in\pi} \left( \lambda_{\mathrm{reg}} \,\mathcal{L}_{\mathrm{reg}}  + \lambda_{\mathrm{cls}} \,\mathcal{L}_{\mathrm{cls}} \right) (\mathcal{T}_{j}, \mathcal{G}_{k}) + \mathcal{L}_{\mathrm{iou}}
\end{equation}
where $\mathcal{L}_{\mathrm{reg}}$ is the $L_1$ loss, $\mathcal{L}_\mathrm{cls}$ represents the focal loss and $\mathcal{L}_\mathrm{iou}$ is the generalized intersection over union loss. 
The weights $\lambda_{reg}$ and $\lambda_{cls}$  balance the contributions of each term.

\begin{table*}[t]
\caption{Comparison between different trackers on nuScenes test set. The upper table shows tracker with different tracking architectures. The lower table focuses on transformer-based trackers. Best value of \textbf{transformer-based} model in bold, second place underlined.}
\centering
\resizebox{\textwidth}{!}{%
\begin{tabular}
{
  l  
  l  
  c  
  c  
  c  
  c  
  c  
  c  
  c  
  c  
  c  
  c  
  c  
  c  
  c  
}
\toprule
\multicolumn{1}{c}{Method} &
\multicolumn{1}{c}{Tracking} &
\multicolumn{1}{c}{Modality} &
\multicolumn{8}{c}{aMOTA $\uparrow$} &
\multicolumn{1}{c}{aMOTP $\downarrow$} &
\multicolumn{1}{c}{Recall $\uparrow$} &
\multicolumn{1}{c}{IDS $\downarrow$} &
\multicolumn{1}{c}{FRAG $\downarrow$} \\
\cmidrule(lr){4-11}
\multicolumn{3}{c}{} &
\multicolumn{1}{c}{Overall} &
\multicolumn{1}{c}{Bic.} &
\multicolumn{1}{c}{Bus} &
\multicolumn{1}{c}{Car} &
\multicolumn{1}{c}{Mot.} &
\multicolumn{1}{c}{Ped.} &
\multicolumn{1}{c}{Tra.} &
\multicolumn{1}{c}{Tru.} &
\multicolumn{1}{c}{} &
\multicolumn{1}{c}{} &
\multicolumn{1}{c}{} &
\multicolumn{1}{c}{} \\
\midrule
SimpleTrack~\cite{simpletrack}     & KF + Hungarian     & L     & 66.8 & 40.7 & 71.5 & 82.3 & 67.4 & 79.6 & 67.3 & 58.7 & 55.0 & 70.3 & 575  & 591 \\
StreamPETR~\cite{wang2023exploringobjectcentrictemporalmodeling}&Greedy matching &C+L&65.3&56.2&60.6&75.1&67.5&72.2&64.6&61.1&56.4&73.3&1037&712\\
FocalFormer3D-F~\cite{focalformer3d}  & KF + Hungarian    & C+L   & 73.9 & 54.1 & 79.2 & 84.0 & 74.4 & 75.2 & 69.6 & 65.2 & 51.4 & 75.9 & 824& 773 \\
TransFusion~\cite{bai2022transfusionrobustlidarcamerafusion}      & Greedy matching        & C+L   & 71.8 & 53.9 & 75.4 & 82.1 & 72.1 & 79.6 & 73.1 & 66.3 & 55.1 & 75.8 &  944  & 673 \\
MSMDFusion~\cite{jiao2023msmdfusionfusinglidarcamera}       & NN/Greedy matching        & C+L   & 74.0 & 57.4 & 76.4 & 82.7 & 74.2 & 80.3 & 72.4 & 66.9 & 54.9 & 76.3 &  1088  & 743 \\
BEVFusion~\cite{10160968bevfusion}       & KF        & C+L   & 74.1 & 56.0 & 77.9 & 84.1 & 73.8 & 77.3 & 74.3 & 69.5 & 40.3 & 77.9 & 506 & 422 \\
Poly-MOT~\cite{10341778polymot}        & Filter-based/NN    & C+L   & 75.4 &58.2& 78.6 &86.5 &81.0 &82.0& 75.1 &66.2& 42.2 & 78.3 &  292  & 297 \\
Fast Poly~\cite{li2024fastpolyfastpolyhedralframework}       & Filter-based/NN   & C+L   & 75.8 & 57.3 & 76.7 & 86.2 & 76.8 & 82.6 & 76.2 & 66.5 & 47.9 & 77.6 & 326 & 270 \\
MCTrack~\cite{Wang2024MCTrackAU}         & Learning-based EKF      & C+L   & 76.3 & 59.2 & 77.1 & 82.9 & 81.7 & 83.2 & 77.3 & 69.2 & 44.5 & 79.1 &  242  & 244 \\
DINO-MOT~\cite{10755965}         & NN        & C+L   & 76.3 & 56.6 & 78.8 & 85.9 & 75.6 & 87.5 & 76.2 & 72.5 & 52.0 & 78.9 &  387  & 470 \\
NeMOT~\cite{wei2025nemot}           & Bayesian NN       & C+L   & 77.9 & 62.5 & 78.4 & 86.7 & 82.5 & 86.1 & 79.3 & 69.8 & 43.6 & 80.5 &  399  & 259 \\
\midrule
MUTR3D~\cite{Mutr3d_camera}& Transformer & C & 27.3 &22.1&23.0&47.4&30.3&228&17.7&19.9&149&44.1&6018&2749\\ 
MotionTrack~\cite{Zhang2023MotionTrackET}     & Transformer      & C+L     & 55.0 &  -& - & - & - &- & - & - & 87.1 & 77.0 & 1321  & 8716 \\
3DMOTFormer~\cite{3dMotformer}      & Transformer       & C+L   & \underline{72.5} & 48.9 & 75.4 & \underline{83.8} & \underline{76.0} & 83.4 & 74.3 & 65.3 & 53.9 & \underline{77.2} &  593  & \textbf{499} \\
LiDAR MOT-DETR~\cite{teye2025lidarmotdetrlidarbasedtwostage}  & Transformer       & L     & 72.4 & \underline{49.0} & \textbf{79.6} & 82.1 & 72.0 & \underline{83.7} & \underline{74.3} & \underline{66.4} & \underline{47.5} & 76.7 &\textbf{404} & 528 \\

\midrule
\addlinespace
\textbf{FutrTrack (Ours)} & Transformer     & C+L &\textbf{74.7}&\textbf{55.8}&\underline{79.0}&\textbf{83.8}&\textbf{76.8}&\textbf{84.2}&\textbf{75.4}&\textbf{67.8}& \textbf{44.9}&\textbf{78.0}&\underline{447}&\underline{523}   \\
\bottomrule
\end{tabular}
}
  
  \label{tab:motmetrics}
\end{table*}

\section{\uppercase{Experiments}}

    

\begin{table*}[t!]
\caption{Comparison of tracker performance with different temporal window sizes of the smoother component using FocalFormer3D camera-LiDAR detections. Temporal window for smoother was set to 15.}
\label{tab:sample_len}
\resizebox{\textwidth}{!}{
\begin{minipage}{.5\linewidth}
  \centering
  \scriptsize
  \setlength{\tabcolsep}{1.5pt}
  \begin{center}
    \begin{tabular}{cccccc}
      \hline
    Total Frames & aMOTA$\uparrow$ & aMOTP$\downarrow$  & IDS$\downarrow$ & FRAG$\downarrow$ &mAP$\uparrow$\\
      \hline
      4 & 0.736 & 0.539 &409 &497 & 0.719\\
      8   &  0.738     & 0.526 & 402 & 478 & 0.731   \\
      12  &  0.741     &  0.499 & 396&465 &0.736         \\
      16 & 0.765      &  0.462 & 376 &  447 & 0.739\\
      
    \hline
    \end{tabular}
    \\(a) offline
  \end{center}
  \label{tab:sample_len_a}
  \end{minipage}

\begin{minipage}{.5\linewidth}
  \centering
  \scriptsize
  \setlength{\tabcolsep}{1.5pt}
  \begin{center}
    \begin{tabular}{cccccc}
      \hline
       Total Frames & aMOTA$\uparrow$ & aMOTP$\downarrow$  & IDS$\downarrow$ & FRAG$\downarrow$ &mAP$\uparrow$\\
      \hline
        4 &0.734  &0.532&419&482&0.708\\
        8 &  0.736 & 0.519 & 411 & 479 & 0.716         \\
      12   & 0.739 &0.493&405&461&0.722  \\ 
      16&  0.752   &0.473 &394&452&0.729 \\
      
    \hline
    \end{tabular}\\
    (b) online
  \end{center}
  \label{tab:sample_len_b}
  \end{minipage}
}

\end{table*}
We evaluate the proposed FutrTrack framework on the nuScenes and KITTI datasets using both single and multi-sensor configurations to validate the core idea of query-based tracking with sensor-fusion modalities. Beyond performance gains, FutrTrack features a modular design that enables seamless integration with different object detectors without requiring architectural modifications. To demonstrate this flexibility, we paired the tracker with two distinct multi-modal detectors and conducted comprehensive experiments across datasets.
The following sections present detailed quantitative results and ablation studies that examine the impact of each sensor modality and architectural component on overall performance. We also analyse the tracker’s detector-agnostic behaviour, showing how different detection inputs influence tracking consistency and accuracy.

\begin{table}
\caption{Performance of FutrTrack when different parts of the input data are missing during inference.}
    \label{defect}
  \scriptsize
  \setlength{\tabcolsep}{1.5pt}

  \begin{center}
    \begin{tabular}{l|c|c|ccccc} 
      \toprule
          Detector &Modality &Defect& aMOTA$\uparrow$   & aMOTP$\downarrow$ &  
          \\
      \midrule
           FocalFormer3D               & C+L &No Front Camera& 0.642 & 0.582                 \\
          FocalFormer3D &C+L &No Back Camera &  0.647     &  0.593        \\
           FocalFormer3D & C+L  &No Back Left Camera           &      0.681            &  0.552                  \\
            FocalFormer3D & C+L  &No Lidar          &     0.568 &0.682          \\
           \bottomrule
    \end{tabular}
    
  \end{center}
\end{table}

\subsection{Implementation Details}

\paragraph{Datasets:}
Evaluation is performed on two widely used benchmarks for 3D perception: the \textbf{nuScenes} dataset and the \textbf{KITTI} dataset.  

The NuScenes dataset is a large-scale benchmark designed for autonomous driving research, particularly perception tasks such as object detection and tracking. It contains data collected from real-world urban environments in Boston and Singapore, and is particularly challenging due to its diverse weather, lighting conditions, and dense traffic scenarios.
Each scene spans 20 seconds and is captured at 2 Hz, providing a rich temporal context for tracking applications.
The dataset includes synchronized multi-modal sensor data:
\begin{itemize}
    \item LiDAR: A roof-mounted, 32-beam spinning LiDAR sensor featuring $360^\circ$° coverage at 20 Hz.
    \item Cameras: Coverage from six cameras covering the full surround view. 
\end{itemize}
It provides $1,\!000$ driving sequences collected in urban environments and contains approximately $1.4$ million annotated object instances across $10$ classes. The 1,000 scenes are divided into three official splits; 700 scenes (~281,000 LiDAR sweeps) for training, 150 scenes (~60,000 LiDAR sweeps) for validation and a test set of 150 scenes (~60,000 LiDAR sweeps).
For evaluation, only the $7$ main class labels are taken into consideration.  

The KITTI dataset, in contrast, focuses on front-view driving scenes and includes synchronized stereo camera for RGB imagery and a Velodyne HDL-64 LiDAR recordings captured in semi-urban and highway settings. KITTI contains around $15,\!000$ labelled 3D bounding boxes over $7,\!000$ frames and remains a standard benchmark for evaluating 3D detection and tracking systems.The KITTI tracking benchmark consists of 21 training sequences and 29 test sequences, totaling over 40 minutes of driving and thousands of annotated frames. 
We focus on the pedestrian and car classes for training and evaluation. The training set includes ground-truth labels for object classes and tracking IDs.

\paragraph{Object Detectors}
As FutrTrack accepts input bounding boxes from an existing object detector, it is evaluated using two state-of-the-art detector architectures.  

First, \textbf{FocalFormer3D} is employed as a transformer-based 3D detector that leverages focal modulation and hierarchical attention to capture both local geometric structures and long-range spatial dependencies in point cloud data.  
Secondly, we utilize \textbf{SparseFusion}, a multi-modal detection framework that performs sparse query-based fusion of LiDAR and camera features. This model enables complementary exploitation of appearance cues from images and precise depth information from LiDAR.  
The output of these detectors serves as inputs to our tracking module, providing high-quality object candidates for temporal association. 

\paragraph{Training:}
The smoother and detector models were implemented and trained from scratch in PyTorch using the MMDetection3D framework. The tracker is trained for 36 epochs on an A100 GPU. We employ the AdamW optimizer with an initial learning rate of 0.001 and apply a step for learning-rate scheduling. For the smoother, the training setup remains the same except that we train for 24 epochs, a batch size of 4, and a reduced initial learning rate of 0.0001.

\begin{figure*}

    \centering
    \includegraphics[width=1.03\textwidth]{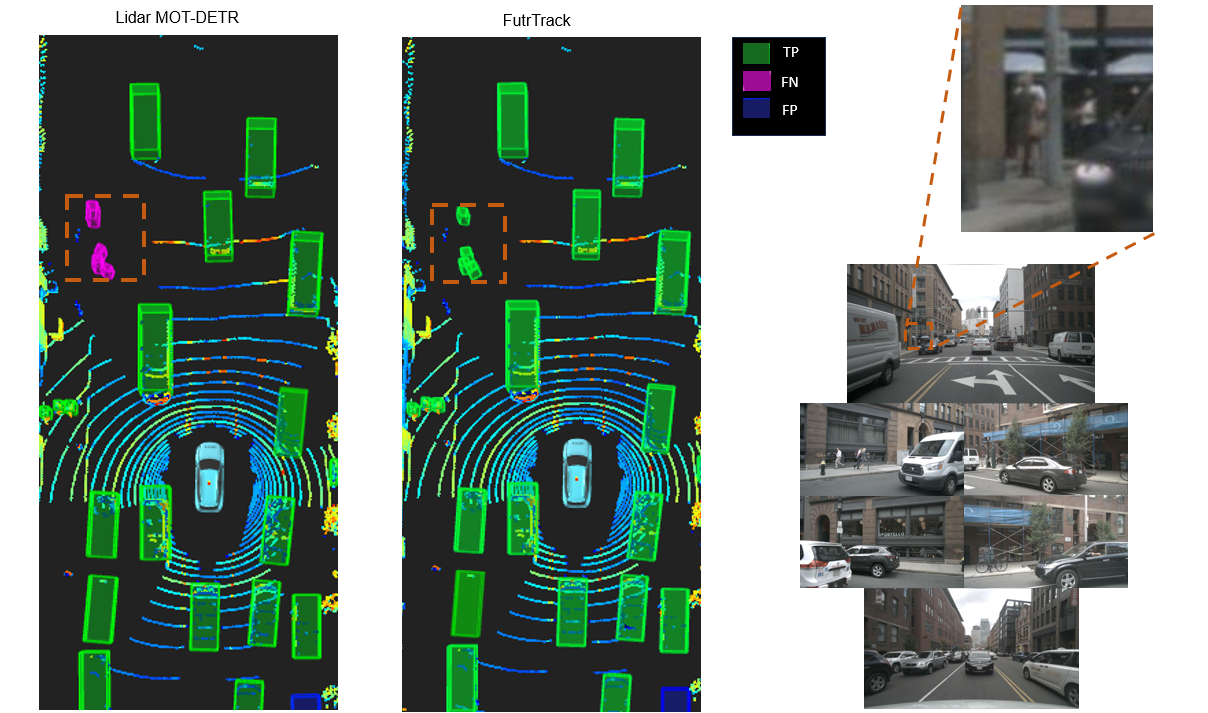}
    \caption{Visual comparison between LiDAR MOT-DETR and FutrTrack. Both models use the same detector and smoother while varying the tracker modality on nuScenes. The front camera captures some pedestrians which are not captured by the LiDAR point cloud. We recommend zooming in to view. (Scene token: $c525507ee2ef4c6d8bb64b0e0cf0dd32$ from nuScenes validation set.)}
    \label{fig:vis}
\end{figure*}

\paragraph{Evaluation Metrics:}
   
       
        
     
The nuScenes tracking benchmark evaluates performance using a suite of metrics to capture both detection quality and temporal consistency. The primary metric is the Average Multi-Object Tracking Accuracy (aMOTA), which extends the concept of MOTA \cite{mota} by averaging over multiple recall thresholds. This measure balances false positives, false negatives, and identity switches, providing a comprehensive indicator of tracking performance. In addition, Average Multi-Object Tracking Precision (aMOTP) is reported to assess the spatial localization accuracy of tracked objects, reflecting how well the estimated boxes align with ground truth. Complementary metrics include IDS/IDSW (Identity Switches), which counts the number of times an object’s identity is incorrectly reassigned, and Fragmentation (FRAG) which counts the number of times a track is broken in different assignments with a video sequence. We also use the nuScenes Detection Score (NDS) to evaluate the smoother's performance.
For KITTI dataset, the Higher Order Tracking Accuracy (HOTA) is used to determine the overall tracking accuracy, as well as the Detection Accuracy (DetA)~\cite{Hota2020IJCV}.

\subsection{Results and Findings} 

We begin by evaluating to what extent fusing image and camera features improves tracking performance. We used LiDAR-MOT-DETR~\cite{teye2025lidarmotdetrlidarbasedtwostage}, retrained with the same camera-LiDAR base detector used in FutrTrack, as a baseline.
We found that replacing LiDAR MOT-DETR's LiDAR-only tracking with FutrTrack's LiDAR-camera tracking improves aMOTA  and aMOTP (table \ref{tracker-robust}) for two different object detectors on both nuScenes and KITTI datasets. For example, in nuScenes, FutrTrack achieves an aMOTA of $75.9$ ($3.1\,pp$ gain) and an aMOTP of $46.0$ ($4.8\,pp$ gain). In particular, we see improved performance in smaller classes, such as bicycles and motorcycles. We also compare the performance of the original LiDAR MOT-DETR (LiDAR-only detector) with that of the retrained LiDAR MOT-DETR (LiDAR + camera detector) and found only a minor performance gain. The increased performance of FutrTrack therefore comes mostly from the camera features added in the tracking itself, not from the performance of the base detector.



\begin{table*}[!htb]
\caption{Overall track metrics on KITTI validation dataset in comparison with other SOTA methods.}
  \label{tab:kitti_data}
  \centering
  \scriptsize
  \setlength{\tabcolsep}{2pt}
  \begin{center}
    \begin{tabular}{lc|c|ccccc}
   
      \toprule
        Name  &&Modality   &HOTA &DetA &IDSW &M0TA \\
    \midrule
        UG3DMOT~\cite{ug3dmot}& & L  &0.808  &0.782 & 13 & 0.866\\
        MCTRACK~\cite{Wang2024MCTrackAU}  &&C+L &0.839  & -& 3 &0.64\\
        BiTrack~\cite{huang2025bitrackbidirectionaloffline3d}  &&C+L &0.845  &0.819 & 13 &0.878\\
        RobMOT~\cite{RobMOT} & & L  &0.863 &- &1 &\textbf{0.915}\\
        CasTrack~\cite{castrack} & & L  &0.932 &- &- &-\\
        PC-TCNN~\cite{pctcnn} & & L  &0.944 &- &3 &0.886\\
        LeGO~\cite{zhang2023lego}& & L &  \textbf{0.952} & -& 1 & 0.90\\
         Lidar MOT-DETR\cite{teye2025lidarmotdetrlidarbasedtwostage}&& L   & 0.852 &0.817 & 12 &0.913 \\
         \midrule
         Ours(FutrTrack)&& C+L   & \underline{0.946} &\textbf{0.821} & 9 & \textbf{0.915}&\\
        
    \bottomrule
    \end{tabular}
     
  \end{center}
  
\end{table*}

\begin{table*}
\caption{Results on nuScenes validation set; with LiDAR-only vs LiDAR-camera fusion as well as with and without smoother. * represents baseline detectors using a Kalman Filter-based tracker (SimpleTrack).}
  \label{tab:mot_val}
  \scriptsize
  \setlength{\tabcolsep}{2.0pt}

  \begin{center}
    \begin{tabular}{l|c|c|c|cccc} 
      \toprule
         Method  & Detector &Modality &Smoother& aMOTA$\uparrow$   & aMOTP$\downarrow$ &  IDS$\downarrow$ & FRAG$\downarrow$ \\
      \toprule
           CenterPoint *~\cite{Centerpoint}&CenterPoint               & L & n/a & 0.637             & 0.606            & -             & -              \\
           Lidar MOT-DETR &CenterPoint &L & \xmark & 0.672 &   0.581       &    456         &   521     \\
           Lidar MOT-DETR & CenterPoint & L& \cmark  & 0.685           & 0.573                  &   438           &  493             \\
           \midrule
        SparseFusion*  ~\cite{xie2023sparsefusion}  &SparseFusion       & C+L & n/a& 0.679            & 0.593 &     603        & 526            \\
      FutrTrack (Ours) &SparseFusion      & C+L&  \xmark & 0.698  &     0.547        &   572          &  512      \\
      FutrTrack (Ours)&SparseFusion         & C+L  & \cmark&    0.732        &  0.525    &  547   &   490           \\
     
      \midrule
     FocalFormer3D*~\cite{focalformer3d} & FocalFormer3D-L        & L &  n/a & 0.705             & - &     -        &        -       \\
       Lidar MOT-DETR & FocalFormer3D-L &L & \xmark   &  0.728            & 0.541 &     419       &        489       \\
        Lidar MOT-DETR& FocalFormer3D-L &L & \cmark   &     0.736         & 0.532 &    392       &        468       \\
        FocalFormer3D*~\cite{focalformer3d} & FocalFormer3D-F        & C+L &  n/a & 0.755             & 0.550 &     653        &        622      \\
         FutrTrack (Ours)&FocalFormer3D-F         & C+L &\xmark &  0.738         & 0.506        &  394     & 476                  \\
        FutrTrack (Ours)&FocalFormer3D-F         & C+L &\cmark &  \textbf{0.759}         & \textbf{0.460}        &  \textbf{340}     & \textbf{429}                 \\
      
      \bottomrule
      
    \end{tabular}
  \end{center}
\end{table*}
\paragraph{Comparison to Other Methods:}
Different trackers rely on different base object detectors which influence the overall tracking performance and are therefore difficult to fairly compare to each other. Even when evaluating FutrTrack with a non-SOTA detector, its performance is competitive with other methods. Table \ref{tab:motmetrics} shows a comparison with other methods as presented by the nuScenes tracking leaderboard. To provide a clearer comparison, we organize recent 3D multi-object tracking methods into categories seen in table \ref{tab:motmetrics} based on their primary tracking strategy. In particular, as shown in the lower part of the table, FutrTrack outperforms all other transformer-based trackers with an aMOTA of 74.7 on the nuScenes test set. Table \ref{tab:kitti_data}) also shows a comparison with other trackers using the KITTI validation dataset with a HOTA of 94.6 and a higher DetA.
Although transformer-based tracking methods do not outperform some classical, Bayesian and geometry-based methods on nuScenes datasets, we expect them to gain performance on larger datasets as has been demonstrated in LiDAR MOT-DETR's~\cite{teye2025lidarmotdetrlidarbasedtwostage} pre-training results.

\paragraph{Effect of Temporal Window in Smoother:}
The performance of the smoother component is affected by the size of the temporal window used as seen in Table \ref{tab:sample_len}. Our findings show that the mAP, aMOTA and aMOTA values improve with larger window sizes. In particular, while we notice improvements in aMOTA, there is a more significant improvement in aMOTP. This shows that a larger temporal window primarily has an effect on the localization and quality of bounding boxes, and less on whether an object is detected in the first place.
\paragraph{Sensitivity to Sensor Malfunction:}
To determine the robustness of our model to sensor defects, we simulate a sensor malfunction or absence during inference and evaluate the model performance in these instances. This malfunction is only simulated at the tracking level and not from the underlying detector since we want evaluate the dependence on the sensor features and tracker's query response when some features are missing. From table \ref{defect}, we notice a more significant drop in performance when the LiDAR sensor is removed compared to when removing one of the camera sensors. However, removing either sensor does not completely degrade the tracking accuracy.
\begin{table*}[t!]
\caption{Smoother results on nuScenes validation using different object camera-LiDAR fusion detectors. * represents results from baseline models.}
  \label{tab:smoother_val}
  \centering
  \scriptsize
  \setlength{\tabcolsep}{1.5pt}
  \begin{center}
    \begin{tabular}{lccccccccc}
      \toprule
      \textbf{Method} & \textbf{Smoother} & \textbf{Detector} &\textbf{Modality} & \textbf{mAP$\uparrow$}& \textbf{mATE$\downarrow$} &  \textbf{mASE$\downarrow$} & \textbf{mAOE$\downarrow$}  & \textbf{mAAE$\downarrow$}   & \textbf{NDS$\uparrow$}  \\
    \midrule
      CenterPoint* \cite{Centerpoint} & N/A &CenterPoint &L&0.564  & -                         &-  & -            & & 0.648      \\
      LiDAR MOT-DETR\cite{teye2025lidarmotdetrlidarbasedtwostage}R & Online~\ &CenterPoint&L& 0.637              &0.255  & \textbf{0.210}             & 0.364            &0.138  & 0.695                \\
     LiDAR MOT-DETR\cite{teye2025lidarmotdetrlidarbasedtwostage} & Offline ~\ &CenterPoint& L&0.671& 0.253 & 0.214               &0.342  & 0.135             & 0.711  \\
     \midrule
      SparseFusion* \cite{xie2023sparsefusion} &N/A      &SparseFusion&L+C& 0.705    & -& -          &    -          &- & 0.728            \\
      Ours(FutrTrack) & Online~\ &SparseFusion&L+C& 0.727              &0.250  & 0.243             & 0.345            &0.131  & 0.739                \\
      Ours(FutrTrack) & Offline ~\ &SparseFusion& L+C&\underline{0.732}& \textbf{0.241} & 0.229               &0.339  & \underline{0.124}             & 0.741  \\
      \midrule
     FocalFormer3D* \cite{focalformer3d} &N/A       &FocalFormer3D&L& 0.664    & -& -          &    -          &- & 0.709            \\          
      
      LiDAR MOT-DETR~\cite{teye2025lidarmotdetrlidarbasedtwostage} & Online &FocalFormer3D&L& 0.672   & 0.249  & 0.245             & 0.331  &  0.125  & 0.725             \\
      LiDAR MOT-DETR~\cite{teye2025lidarmotdetrlidarbasedtwostage} & Offline~&FocalFormer3D&L & 0.683   & 0.251  & 0.242             & 0.337             & 0.133  & 0.717 \\
        
      FocalFormer3D* \cite{focalformer3d} &N/A        &FocalFormer3D&L+C& 0.705    & 0.275& 0.255          &    0.278          &0.185 & 0.731           \\

      Ours(FutrTrack) & Online&FocalFormer3D&L+C& 0.728   & 0.254  & 0.231             & \underline{0.226}             & 0.128  & \underline{0.743}            \\
      Ours(FutrTrack) & Offline~&FocalFormer3D&L+C & \textbf{0.735}   & \underline{0.248}  & \underline{0.213}             & \textbf{0.223}             & \textbf{0.121}  & \textbf{0.749} \\

      \bottomrule
      
    \end{tabular}
  \end{center}
  
\end{table*}

\paragraph{Effect of Object Detector Used:}
To assess the generality of our tracking framework, two different state-of-the-art 3D object detectors were used and independently evaluated. By evaluating our method with each detector separately, its robustness was analysed across purely LiDAR-based perception and multi-modal fusion-based perception. 
With SparseFusion~\cite{xie2023sparsefusion} detector, our model achieves a 1.9\,pp and 5.3\,pp improvement in aMOTA when compared to SimpleTrack without and with the smoother model, respectively (table \ref{tab:mot_val}).   
When using FocalFormer3D as the base detector the improvement in aMOTA is more modest, and instead we see a gain of 3.4\,pp and 9\,pp in aMOTP (without and with smoother, respectively) compared to SimpleTrack, corresponding to a higher quality of the resulting bounding boxes. The smoother component has two variations: online and offline (forward-looking), see table \ref{tab:smoother_val} for an evaluation and comparison. For a fairer comparison with other tracking methods, all results in this paper use the online variant of the smoother, unless explicitly stated otherwise.



\section{\uppercase{Conclusion}}

This work shows that integrating camera features with LiDAR significantly improves tracking performance, especially in dense urban environments. FutrTrack achieves this through a unified camera–LiDAR BEV representation that allows the transformer architecture to leverage complementary spatial and semantic cues. Its modular design supports off-the-shelf detectors without architectural changes, ensuring adaptability to evolving perception pipelines. Experiments on nuScenes and KITTI demonstrate strong generalization across datasets and detectors, with FutrTrack outperforming previous transformer-based methods. Although geometry-based trackers still lead in some benchmarks, we expect multi-modal transformer approaches like FutrTrack to surpass them as larger and more diverse training data become available. Overall, FutrTrack sets a new direction for unified camera–LiDAR transformer tracking and offers a promising path toward scalable, real-world multi-object tracking.

\bibliographystyle{apalike}
{\small
\bibliography{example}}
\clearpage

\end{document}